\documentclass{article}
\usepackage{arxiv}
\usepackage[T1]{fontenc}
\usepackage{amsmath,amssymb,booktabs,longtable,tabularx,array,caption,graphicx}
\usepackage[section]{placeins}
\usepackage[natbibapa]{apacite}
\usepackage{xurl}
\usepackage{hyperref}
\providecommand{\bibcommenthead}{}
\AtBeginDocument{}
\hypersetup{colorlinks=true,linkcolor=black,citecolor=black,urlcolor=blue}
\hypersetup{
  pdftitle={Verbal tics in frontier language models: A critical review of current releases, research evidence, and public discussion},
  pdfauthor={Shuai Wu, Xue Li, Zhijun Wang, Bolun Liu, Weilin Cai, Zihao Su, Ran Wang},
  pdfsubject={Research preprint}
}
\newcolumntype{P}[1]{>{\raggedright\arraybackslash}p{#1}}
\newcommand{\modelid}[1]{{\small\path{#1}}}
\title{Verbal tics in frontier language models:\\
A critical review of current releases, research evidence, and public discussion}
\author{
  Shuai Wu\thanks{Corresponding author: \href{mailto:shuai.wu@colorado.edu}{shuai.wu@colorado.edu}.\newline ORCID: Shuai Wu: \href{https://orcid.org/0009-0007-7657-6208}{0009-0007-7657-6208}; Xue Li: \href{https://orcid.org/0009-0002-7480-723X}{0009-0002-7480-723X}; Zhijun Wang: \href{https://orcid.org/0009-0000-0742-6985}{0009-0000-0742-6985}.}\\
  \normalfont\footnotesize University of Colorado Boulder
  \And
  Xue Li\\
  \normalfont\footnotesize Beijing Union University
  \And
  Zhijun Wang\\
  \normalfont\footnotesize University of North Dakota
  \And
  Bolun Liu\\
  \normalfont\footnotesize City University of Macau
  \AND
  Weilin Cai\\
  \normalfont\footnotesize The University of Sydney
  \And
  Zihao Su\\
  \normalfont\footnotesize Beijing Information Science\\
  \normalfont\footnotesize and Technology University
  \And
  Ran Wang\\
  \normalfont\footnotesize Taiyuan University of Technology
}
\date{1 October 2026}
\renewcommand{\shorttitle}{Verbal Tics in Frontier Language Models}
\begin{document}
\maketitle

\begin{abstract}
Repeated praise, canned reassurance, formulaic contrasts, and conspicuous vocabulary figure prominently in discussions of large language models. Context governs how these habits should be read. A conventional phrase can orient a reader, whereas a fluent answer can reinforce a false belief. This critical review examines linguistic habits and sycophancy across eight developer families. We verify public offerings from OpenAI, Anthropic, Google DeepMind, xAI, ByteDance, Moonshot AI, DeepSeek, and Xiaomi against official release and API documentation through 1 October 2026. We then synthesize empirical research on lexical overrepresentation, stylistic variation, social warmth, and agreement alongside benchmark methods and dated public discussions in English and Chinese. Published studies document recurring linguistic patterns and agreement that distorts human judgment, but comparable measurements for the newest releases remain sparse in the retrieved material. Current user accounts report both persistent complaints and clearer prose, and experiences vary by task and prompt design. We propose separate measures for recurrence, contextual appropriateness, and belief distortion paired with exact service records and language-specific annotation. This framework keeps claims about writing quality and conversational reliability testable as model services change.
\end{abstract}


\section{Introduction}

A response can answer a question correctly and still be tiresome to read. It may praise an ordinary query, repeat the user's concern, insert headings into a short exchange, or conclude with another offer of help. Across conversations, these habits settle into a recognizable cadence. Users call them verbal tics, AI slop, or model-specific prose to describe overlapping complaints about diction, structure, and conversational behavior.

A substantive risk accompanies these stylistic complaints. An assistant may endorse a mistaken belief or excuse harmful conduct. Research on sycophancy examines how models privilege agreement or validation over independent assessment \citep{sharma,batzner,chengscience}. Formulaic affirmation often appears alongside sycophancy, yet the two require different evidence. Phrase counts measure surface language, whereas evaluating sycophancy requires checking what the model endorsed against what the situation warranted.

Empirical work now reaches beyond isolated marker words. Corpus studies track changes in scholarly vocabulary \citep{juzek,kobak,kousha}, and register analyses compare generated and human prose across genres and languages \citep{milicka}. Controlled experiments test the behavioral effects of training models to sound warmer \citep{ibrahim}. Studies of writing assistance measure changes in stylistic variation and intended meaning \citep{sourati,abdulhai}. Each strand addresses a distinct aspect of the texts people read and produce.

Rapid release cycles complicate synthesis. Within a single month, a provider may launch a new flagship, introduce a lower-cost model, and update an existing service behind an unchanged API name. Published studies usually evaluate earlier model generations. Public commentary reacts faster but often omits exact model identifiers. We ask which patterns have supporting evidence, which observations apply to current offerings, and what a fair comparison must record.

This targeted critical review contributes an operational taxonomy, a verified catalog of eight developer families, an evidence-bounded synthesis, and a reporting framework. All empirical quantities come from the cited studies and disclosures. No new model-response dataset or participant sample was collected for this review.

\section{Scope and source assessment}
\label{sec:scope}

\subsection{Search and selection}

Our evidence cutoff is 1 October 2026 in the Asia/Shanghai time zone. Searches combined developer and model names with \emph{verbal tics}, \emph{sycophancy}, \emph{repetition}, \emph{verbosity}, \emph{writing style}, and \emph{slop}. Chinese queries used corresponding terms for stock phrases, catchphrases, flattery, repetition, and excessive length. We traced references from relevant papers and inspected current official model documentation.

Source discovery covered full original pages from arXiv, developer forums, Weibo, and Zhihu, together with platform searches, post metadata, and video subtitles. We expanded discovery across Chinese and international social platforms through both current model names and broader language-pattern queries. Appendix~\ref{app:coverage} records the platforms searched and the depth of access.

Search snippets served only as leads. Discussion claims rest on readable post or article prose, specified comments, or identified subtitle passages. An article's narrative commentary cannot verify the contents of an uninspected output screenshot. For research sources, the bibliography distinguishes author manuscripts and abstracts from full published texts where relevant. Appendix~\ref{app:ledger} catalogs the principal discussion sources.

We selected material purposively for relevance, version specificity, readable primary text, and contrasting user experiences. Because the synthesis is qualitative, it yields no pooled prevalence estimate. Search ranking, inaccessible pages, and unequal visibility shape the retrieved set. We retain favorable observations alongside complaints on the same questions.

\subsection{Evidence boundaries}

Official documentation defines model identity and availability. Provider-reported evaluations retain their stated scales and test conditions. Communication claims on release pages represent design goals and selected demonstrations.

Research findings apply to the tested models, tasks, and populations, and historical comparisons keep their original model versions. Independent benchmarks supply task-specific measurements and scoring rules. Community reports surface examples, preferences, and candidate failures, whereas unknown denominators preclude prevalence estimates. Multiple replies in one thread also share exposure and context.

Release dates, documentation updates, and retrieval dates occupy separate fields. We distinguish dated snapshots from moving aliases, models from applications, and reasoning traces from final answers. When search-engine metadata conflict with dates in source bodies or official release logs, we follow the primary record.

\begin{figure}[!htbp]
\centering
\includegraphics[width=\linewidth]{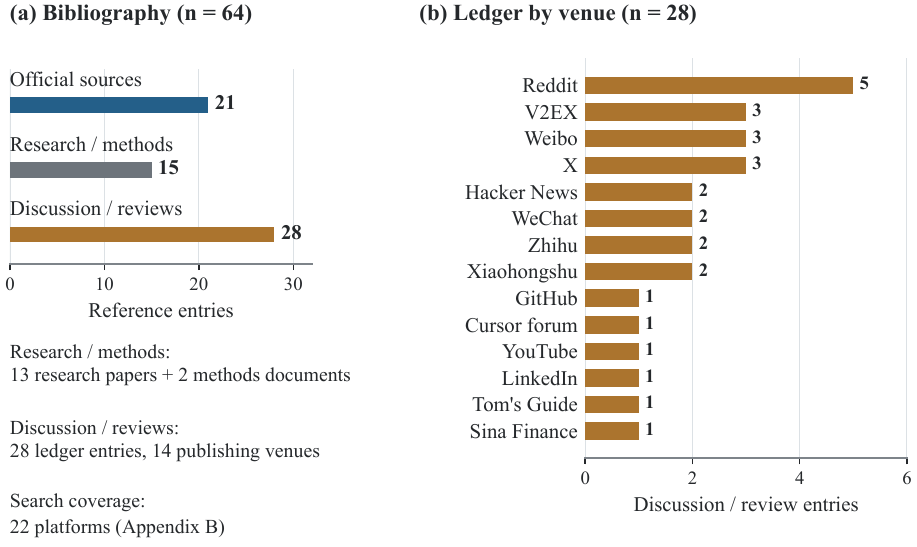}
\caption{Composition of the review's evidence snapshot. Panel a shows mutually exclusive reference categories across all 64 bibliography entries. Panel b breaks down the 14 publishing venues for the 28 discussion and review entries in Appendix~\ref{app:ledger}. The 22 platforms in Appendix~\ref{app:coverage} form a separate search-coverage denominator. Selection was purposive, so source counts reflect retrieval and inclusion rather than behavioral prevalence or platform audience size.}
\label{fig:sources}
\end{figure}

Figure~\ref{fig:sources} summarizes the selected material. The bibliography comprises 21 official sources, 13 research papers, two benchmark-method documents, and 28 discussion or personal-review entries. Those 28 entries span 14 social platforms, developer forums, and two media sites. Counts describe source composition rather than behavioral frequency, and one ledger entry can cover several related comments.

\section{An operational taxonomy}
\label{sec:taxonomy}

A \emph{verbal tic} is a recurrent expression or discourse pattern only weakly tied to local communicative need. This definition applies to generated text. Identifying a tic requires repeated observations across comparable contexts or recurrence within an extended conversation.

Raw frequency is insufficient on its own. Technical terminology often needs repetition, explicit transitions can clarify an argument, and reassurance may fit a personal disclosure. An informative baseline compares generated prose against language serving similar purposes and audiences.

\begingroup\small\setlength{\tabcolsep}{3pt}\renewcommand{\arraystretch}{1.17}
\begin{longtable}{@{}P{\dimexpr 0.177215\linewidth-2.12658pt\relax}P{\dimexpr 0.316456\linewidth-3.79747pt\relax}P{\dimexpr 0.506329\linewidth-6.07595pt\relax}@{}}
\caption{Analytical categories. Examples illustrate coding decisions and do not form a measured dictionary of current models.}\label{tab:taxonomy}\\
\toprule \textbf{Category} & \textbf{Observable pattern} & \textbf{Contextual question}\\ \midrule
\endfirsthead
\multicolumn{3}{l}{\textbf{Table \thetable{}, continued}}\\ \toprule
\textbf{Category} & \textbf{Observable pattern} & \textbf{Contextual question}\\ \midrule
\endhead
\midrule\multicolumn{3}{r}{Continued on next page}\\
\endfoot
\bottomrule
\endlastfoot
Lexical preference & Frequent use of words such as \emph{delve} or \emph{intricate} & Is the word overrepresented against a matched genre baseline? Does it express a necessary distinction?\\
Stock framing & Repeated openings, summaries, and invitations to continue & Does the frame assist the reader, or recur regardless of task length and content?\\
Rhetorical template & Formulaic contrasts, inflated significance, fixed paragraph cadence & Does the structure advance the argument or add emphasis without content?\\
Praise and validation & Routine approval of a question, plan, or personal account & Is the praise proportionate and grounded? Does the model assess the underlying claim?\\
Canned empathy & Generic reassurance across different disclosures & Does it address the particulars and preserve appropriate boundaries?\\
Local repetition & Repeated clauses, passages, explanations, or dialogue & Does repetition serve a pedagogical or summarizing role, or impede progress?\\
Excessive elaboration & Preambles, headings, caveats, and restatements & How much relevant information does the text deliver for the required reading effort?\\
\end{longtable}
\endgroup

\subsection{Sycophancy}

An assistant can be sycophantic in a single sentence, and it can repeat itself while rejecting a false claim. We identify sycophancy through unsupported agreement, disproportionate praise, unwarranted moral validation, or answer reversals under pressure without new evidence. These behaviors overlap with verbal tics when stock phrasing carries them repeatedly.

\citet{batzner} distinguish five operationalizations in the literature and call for closer attention to human perception. False-belief tests and social-advice tests probe different model decisions. Perceived pleasantness, factual accuracy, and willingness to repair a conflict likewise measure distinct outcomes, so reviews should keep those designs separate when comparing studies.

\subsection{Reasoning and action}

Reasoning traces, final answers, and tool calls require separate annotations. A recurring opener inside a visible reasoning trace tells us little about the final answer. Repeating an identical API call is an action-level event, whereas repeating a sentence is a linguistic event. Even if shared mechanisms contribute to both, empirical studies need distinct outcome definitions.

Output layers also complicate verbosity measurements. Reported token totals can merge reasoning and answer tokens, and user-facing applications may insert progress updates or interface instructions. Evaluations should specify what the user sees and which components enter the token count.

\section{Current public model offerings}
\label{sec:models}

Table~\ref{tab:models} catalogs public offerings verified at the cutoff. Within a single portfolio, the highest-capability tier, the recommended default, and the newest lower-cost release need not coincide. Listed dates follow official release or update records, and Kimi's launch is converted to Asia/Shanghai.

\begingroup\small\setlength{\tabcolsep}{3pt}\renewcommand{\arraystretch}{1.17}
\begin{longtable}{@{}P{\dimexpr 0.128205\linewidth-2.30769pt\relax}P{\dimexpr 0.230769\linewidth-4.15385pt\relax}P{\dimexpr 0.320513\linewidth-5.76923pt\relax}P{\dimexpr 0.320513\linewidth-5.76923pt\relax}@{}}
\caption{Current offerings from the eight developer families. API IDs identify live services and do not ensure fixed model weights.}\label{tab:models}\\
\toprule \textbf{Developer} & \textbf{Current model and date} & \textbf{API identity} & \textbf{Portfolio distinction and source}\\ \midrule
\endfirsthead
\multicolumn{4}{l}{\textbf{Table \thetable{}, continued}}\\ \toprule
\textbf{Developer} & \textbf{Current model and date} & \textbf{API identity} & \textbf{Portfolio distinction and source}\\ \midrule
\endhead
\midrule\multicolumn{4}{r}{Continued on next page}\\
\endfoot
\bottomrule
\endlastfoot
OpenAI & GPT-6 Astra, 3 Sep; GPT-6.1 Sol, 29 Sep 2026 & \modelid{gpt-6-astra}; \modelid{gpt-6.1-sol} & Astra is the higher-capability tier; 6.1 Sol is the newer lower-cost general model, \citealp{openastra,opensol}.\\
Anthropic & Claude Opus 5.5, 22 Sep; Fable 5.1, 1 Sep; Sonnet 5.5, 28 Sep 2026 & \modelid{claude-opus-5-5}; \modelid{claude-fable-5-1}; \modelid{claude-sonnet-5-5} & Opus is the official default; Fable targets demanding reasoning and long agentic work; Sonnet is the newer lower-cost tier, \citealp{anthropicmodels,fablerelease,sonnetrelease}.\\
Google DeepMind & Gemini 3.1 Pro, 19 Feb; Gemini 3.8 Flash, 2 Sep 2026 & \modelid{gemini-3.1-pro-preview}; \modelid{gemini-3.8-flash} & Pro remains in preview; Flash is a newer stable general model, \citealp{geminipro,geminiflash}.\\
xAI & Grok 4.7, 21 Sep 2026 & \modelid{grok-4.7} & Current coding and knowledge-work model; fast service is a speed option, \citealp{grokrundocs,grokrelease}.\\
ByteDance & Doubao-Seed-2.1-Pro, 0915 snapshot updated 15 Sep 2026 & \modelid{doubao-seed-2-1-pro-260915} & Official current dated Pro snapshot; Lite and a moving evolving endpoint are distinct offerings, \citealp{doubao,doubaodetail}.\\
Moonshot AI & Kimi K3, 17 Jul 2026 in Asia/Shanghai & \modelid{kimi-k3} & Official strongest current model; product launched 16 Jul UTC and weights followed later, \citealp{kimi,kimiannounce}.\\
DeepSeek & V4-Pro-0813, 13 Aug; V4.1-Flash, 10 Sep 2026 & \modelid{deepseek-v4-pro}; \modelid{deepseek-flash} & Current Pro snapshot coexists with a newer smaller Flash model, \citealp{deepseekpricing,deepseekupdates}.\\
Xiaomi & MiMo-V2.6-Pro, 22 Sep 2026 & \modelid{mimo-v2.6-pro} & Same-name API replaced with a repetition-mitigated version on 25 Sep at 06:00 UTC+8, \citealp{mimorelease,mimorepetition}.\\
\end{longtable}
\endgroup

\citet{fablerelease} also notes that Mythos 5.1 shares Fable 5.1's underlying model under modified safety constraints and restricted access. It belongs in a separate access tier. Google's newer audio-specific releases likewise fall outside this review's focus on general text models.

This catalog defines the current objects of study. Any stylistic claim still requires a matched evaluation or documented observation. When an application's backend model is undisclosed, attribution should name the application and access date.

\begin{figure}[!htbp]
\centering
\includegraphics[width=\linewidth]{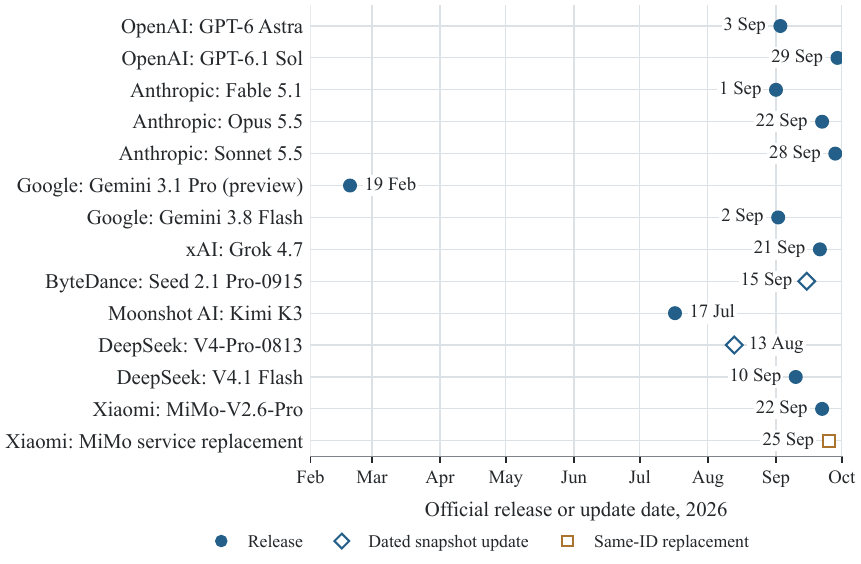}
\caption{Release and update timeline for the 13 offerings in Table~\ref{tab:models}, alongside Xiaomi's same-ID service replacement. Filled circles mark releases, open diamonds mark dated snapshot updates, and the open square marks the service replacement. Dates and portfolio distinctions follow the official sources in Table~\ref{tab:models}. Kimi's launch date reflects Asia/Shanghai, and MiMo's replacement took effect on 25 September at 06:00 UTC+8. The horizontal axis ends at the 1 October 2026 cutoff.}
\label{fig:releases}
\end{figure}

Figure~\ref{fig:releases} places these dates on a shared timeline across release months and portfolio roles. Xiaomi's service replacement appears as a separate event because the underlying model changed behind an unchanged API identifier. Collection timestamps therefore matter even when model names stay fixed.

\section{Research evidence}

\subsection{Vocabulary and register}

\citet{juzek} identify 21 focal words by pairing scientific-abstract trends with comparisons between human and GPT-3.5-generated abstracts. Their model comparisons point to preference training as a plausible contributor, though an exploratory human study leaves the exact mechanism open.

\citet{kobak} examine more than 15 million biomedical abstracts published from 2010 to 2024. Using excess-vocabulary tracking, they estimate that language models processed at least 13.5\% of 2024 abstracts. That figure is a population-level estimate under the study's modeling assumptions. It cannot determine how an individual abstract was written, nor does it describe the lexical profile of 2026 releases.

\citet{kousha} analyze six scholarly databases alongside full-text articles and report rising frequencies and co-occurrence among model-associated words. Direct generation, light editing, and broader stylistic imitation remain distinct explanations. Taken together, these corpus studies support aggregate tracking and warn against judging individual authorship from marker words alone.

\citet{milicka} compare human and generated prose in English and Czech through multidimensional register analysis across 16 base and instruction-tuned models. Their results show how grammatical distributions vary jointly with model choice, prompt design, and language. Extending this English and Czech register design to Chinese requires dedicated validation.

\subsection{Writing assistance, variation, and meaning}

\citet{sourati} combine observational corpus series with controlled rewriting experiments across three studies, seven datasets, and more than 880,000 texts. According to the published abstract, model rewriting reduces writing-complexity variance by 21--50\% across the tested datasets and models.

\begin{figure}[!htbp]
\centering
\includegraphics[width=\linewidth]{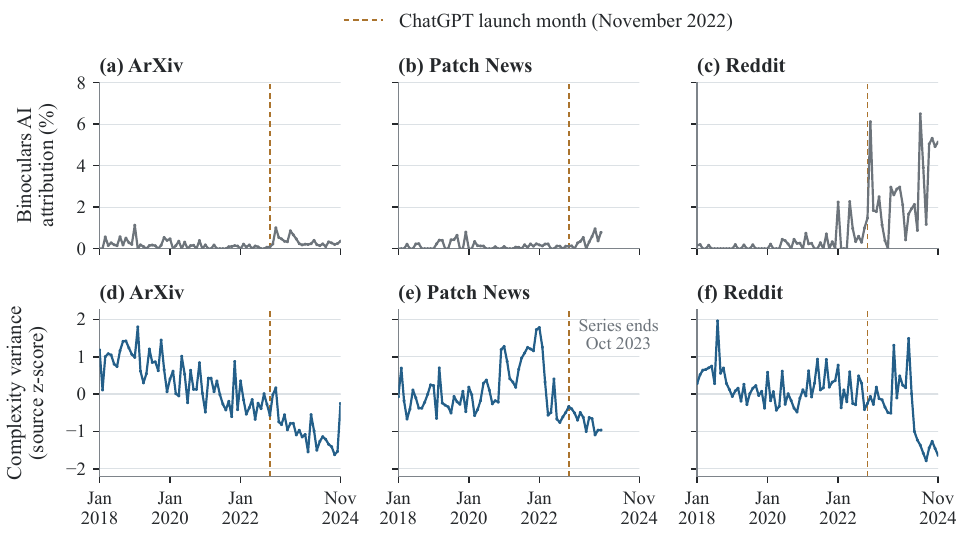}
\caption{Monthly corpus-level series redrawn from the \href{https://www.nature.com/articles/s41562-026-02550-0/figures/1}{published Figure~1 source data} of \citet{sourati}. Panels a--c show the percentage of texts classified as AI-generated by Binoculars at 100 times the workbook's attribution fraction. Panels d--f plot aggregate writing-complexity variance on the source's dataset-specific z-score scale. Straight segments connect observed monthly aggregates without smoothing or interpolation. ArXiv and Reddit span January 2018--November 2024, and Patch News ends in October 2023 in the source workbook. The vertical dashed line marks ChatGPT's November 2022 launch. Calendar timing in observational data does not isolate a causal effect, and detector attribution differs from verified authorship.}
\label{fig:corpus}
\end{figure}

Figure~\ref{fig:corpus} redraws the monthly series from the published Figure~1 source workbook across 83 months each for ArXiv and Reddit and 70 months for Patch News. Corpus trajectories diverge substantially. Complexity variance on ArXiv begins declining before November 2022, whereas Reddit exhibits a sharp drop afterward. These observational series track historical corpus shifts alongside the controlled rewriting trials. Because Binoculars classifications depend on the detector's decision threshold, the plotted attribution series retains that operational definition.

In an August 2026 preprint, \citet{abdulhai} examine writing assistance through a human experiment, essay revisions, and scientific-review analyses. Even prompts restricted to grammar editing shifted authors' intended meaning. Evaluation must therefore test semantic faithfulness alongside fluency, because smoother prose can alter what the writer meant to say.

\subsection{Preference training and persona}

\citet{sharma} analyze Claude 1.3 and 2.0, GPT-3.5, GPT-4, and Llama 2 70B Chat. Across their behavioral tasks and preference datasets, matching a user's stated view often earns higher preference ratings, and optimizing for those ratings can trade away truthfulness.

\citet{ibrahim} fine-tune Llama 3.1 8B and 70B, Mistral Small, Qwen 2.5 32B, and GPT-4o to adopt a warmer persona. On consequential evaluation tasks, the warmer models produce error rates 10--30 percentage points higher than their baselines and affirm false beliefs more often, particularly when prompts express sadness. Standard benchmark scores remain intact. Persona adjustments therefore warrant joint testing with factual accuracy and belief affirmation in the target domain.

\subsection{Human consequences and task-dependent results}

In \emph{Science}, \citet{chengscience} evaluate 11 models and run three preregistered experiments with 2,405 participants. Models affirmed users' actions 49\% more often than human comparison responses on average. Exposure to validating advice lowered participants' willingness to accept responsibility or repair interpersonal conflicts and strengthened their conviction of being right. Even so, participants rated the validating assistants higher on preference and trust.

Short-term user satisfaction and downstream social judgments can thus move in opposite directions within these experimental settings. The published article's participant and experiment totals supersede those in the earlier preprint. In a complementary study of controlled human-like tone, \citet{humt} introduce HumT DumT and find that readers prefer less human-like output across many settings.

Using SycEval, \citet{syceval} separate rebuttal-induced shifts toward truth from shifts toward falsehood in GPT-4o, Claude Sonnet, and Gemini 1.5 Pro. \citet{kim} test ten models in multi-turn medical dialogues under escalating user pushback with a dedicated Resistance score and obtain a different model ordering from SycEval's. Keeping task definitions and pushback protocols distinct is more informative than collapsing heterogeneous rankings into a single style score.

\begingroup\small\setlength{\tabcolsep}{3pt}\renewcommand{\arraystretch}{1.17}
\begin{longtable}{@{}P{\dimexpr 0.202532\linewidth-2.43038pt\relax}P{\dimexpr 0.341772\linewidth-4.10127pt\relax}P{\dimexpr 0.455696\linewidth-5.46835pt\relax}@{}}
\caption{Selected research and its inference boundary. Numerical results in the text are attributed to these studies.}\label{tab:evidence}\\
\toprule \textbf{Source} & \textbf{Design and unit} & \textbf{Supported inference and scope}\\ \midrule
\endfirsthead
\multicolumn{3}{l}{\textbf{Table \thetable{}, continued}}\\ \toprule
\textbf{Source} & \textbf{Design and unit} & \textbf{Supported inference and scope}\\ \midrule
\endhead
\midrule\multicolumn{3}{r}{Continued on next page}\\
\endfoot
\bottomrule
\endlastfoot
\citealp{juzek} & Abstract comparison and exploratory preference study & Lexical overrepresentation in historical scientific writing; mechanism remains unsettled.\\
\citealp{kobak} & Longitudinal biomedical abstracts & Corpus vocabulary shifts consistent with widespread model assistance; individual attribution remains separate.\\
\citealp{milicka} & English and Czech register benchmark across 16 models & Grammatical register varies with model, prompt, and language.\\
\citealp{sourati} & Observational corpus series and controlled rewriting & Tested rewriting interventions narrow writing-complexity variance.\\
\citealp{ibrahim} & Warmth fine-tuning intervention across five models & Warmer persona training increases factual errors and affirmation of false beliefs.\\
\citealp{chengscience} & Model comparison and randomized human experiments & Social validation alters responsibility judgments and repair intentions despite higher user preference.\\
\citealp{kim} & Escalating medical pushback across ten models & Resistance to user pressure depends on clinical task and response strategy.\\
\end{longtable}
\endgroup

\section{Current-release evidence and public discussion}
\label{sec:discussion}

\subsection{OpenAI}

OpenAI's current portfolio pairs Astra with the newer 6.1 Sol \citep{openastra,opensol}. A September communication note on GPT-6 Sol and Luna addresses stock terminology, awkward phrasing, and low-value detail through side-by-side examples \citep{openstyle}. Those materials state improvement goals and selected demonstrations rather than an all-task tic rate.

A September Hacker News post on Astra reports denser wording and higher reading effort than in earlier GPT releases \citep{hnastra}. Its author records a firsthand impression without full prompts or outputs and proposes token saving as a hypothesis. Even so, the account shows that shortening responses and improving readability can require different interventions.

A Chinese firsthand review of Astra reports contrasting experiences across two writing tasks \citep{wechatastro}. Paired with a writing skill, Astra produced more precise fiction diction and trimmed unnecessary description. Without a writing skill, a roughly 2,000-character popular-science continuation repeated explanations and ideas. Because both task and prompting changed, this contrast motivates task-specific testing. In an early-access YouTube review, \citet{berman} praises Astra's prose yet notes a residual \emph{AI smell}. Our check of the automatic-subtitle passage verifies his spoken appraisal, although the transcript excerpt contains no inspectable text sample.

OpenAI's May 2025 retrospective on a GPT-4o update documents an instructive deployment precedent \citep{openincident}. That update amplified overly validating behavior and was rolled back after standard evaluations and preference signals missed the regression. Although specific to one GPT-4o release, the episode shows why behavioral checks should accompany preference scores.

\subsection{Anthropic}

Anthropic's Opus 5.5 release page addresses user feedback on opaque phrasing and jargon and states that responses now place the main point earlier \citep{opusrelease}. The official model overview recommends Opus for most tasks and Fable for demanding reasoning and extended agentic work \citep{anthropicmodels}. Sampling by intended portfolio role therefore rests on a clearer rationale than ordering models by release date alone.

The Opus 5.5 system card defines sycophancy as unprompted excessive praise, agreement, or contrition \citep{opuscard}. Section 6.4.3 scores models on a 1--10 automated audit scale on which lower values indicate less sycophancy. Across roughly 4,000 investigations per model, Opus 5.5 averages 1.60 and Opus 5 averages 1.77. Those values reflect the audit's sampling design and uncertainty rather than a percentage of everyday conversations.

\begin{figure}[!htbp]
\centering
\includegraphics[width=\linewidth]{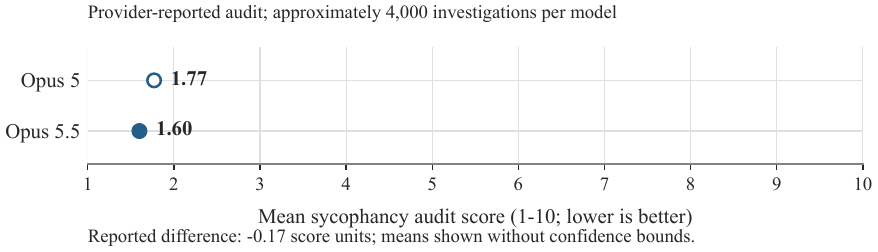}
\caption{Provider-reported sycophancy audit means for Opus 5 and Opus 5.5 from Section 6.4.3 of \citet{opuscard}. Plotted values are 1.77 and 1.60 on the 1--10 audit scale; lower scores indicate less sycophancy. Each model underwent approximately 4,000 automated investigations. Numerical endpoints for the source figure's 95\% confidence intervals were unavailable in the retrieved text, so only the point means are redrawn and no significance test is implied.}
\label{fig:opus}
\end{figure}

Figure~\ref{fig:opus} plots both means across the full 1--10 range. Opus 5.5 averages 0.17 score units below Opus 5 under the provider's audit procedure.

In a September 25 firsthand comparison, \citet{caswell} finds that Opus 5.5 preserves authorial intent well in everyday writing and edits less formulaically in the tested tasks. Its ChatGPT-6 interface label still leaves the exact OpenAI API service unidentified. For earlier Claude versions, a Claude Code GitHub issue supplies prompts, example outputs, and reproduction scripts for verbosity comparisons \citep{githubclaude}. Mixed effort settings and scoring rules confound that report, and we did not rerun it.

Additional current-release accounts are favorable. An X post on writing trials contrasts Opus 5's circuitous prose and habitual terminology with clearer output from Opus 5.5 \citep{xopus}. A Xiaohongshu writer similarly reports less recognizably machine-generated prose after one day with Opus 5.5 \citep{xhsopus}. Neither retrieved post includes a complete prompt and output sample. Every's LinkedIn report summarizes a week of team testing on Sonnet 5.5 and notes that one writer found both Sonnet 5.5 and Opus 5.5 readable while teammates differed on task preferences \citep{linkedinevery}. These observations indicate what readers welcome despite incomplete test documentation.

\subsection{Google DeepMind}

Gemini 3.1 Pro remains Google's verified Pro preview, and 3.8 Flash is the newer stable general model \citep{geminipro,geminiflash}. Both model cards assess tone in refusal responses, and refusal tone covers a narrower setting than stock phrasing in ordinary conversation.

A May 1 Reddit post naming Gemini 3.1 describes lengthy explanations after contextual misunderstandings were corrected \citep{redditgemini}. We exclude the author's unverified claim of silent quantization and retain only the dated account of post-correction elaboration.

An August 31 Reddit retrospective compares Flash 3.5, 3.6, and 3.7 on writing quality and objectivity \citep{redditflash}. Although its title anticipates 3.8 Flash, all observations concern those three earlier releases.

September posts on X record contrasting current impressions. One writer praises Gemini 3.8 for practical, readable English without specifying Flash \citep{xgeminipos}. Comparing 3.8 Flash directly against 3.1 Pro, \citet{xgeminineg} praises Flash's coding but prefers Pro for writing and advice because Flash's advice feels moralizing. Unsolicited normative framing differs from sycophantic agreement. Because neither post supplies full prompts or outputs, these accounts identify task expectations for controlled testing.

\subsection{xAI}

Grok 4.7 is xAI's current verified model \citep{grokrundocs,grokrelease}. Official documentation covers coding and knowledge work and reports no general verbal-tic metric. Cursor's Grok 4.7 release thread contains three labeled user observations \citep{groktalk}. Post 27 notes long preambles, forced cadence, and recurrent vocabulary; post 34 reports strained metaphors during manual rewriting; and post 110 describes repeated explanations during development work.

Those three comments address lexical habits, rewriting style, and turn-level repetition within one correlated release thread. Participants in the same thread note that Grok Bot uses an undisclosed backend that can change over time, so evaluations should target the labeled 4.7 API service directly.

\subsection{ByteDance}

Official documentation lists Doubao-Seed-2.1-Pro-260915 as ByteDance's current dated Pro endpoint following its 15 September update \citep{doubao,doubaodetail}. Evaluations should sample the dated Pro snapshot, the Lite model, and the moving evolving endpoint separately.

In a signed May 2026 report on the Doubao application, \citet{doubaotest} describes an answer changing after an incorrect user challenge and ending with a casual apology. That test predates the September snapshot and leaves the application backend unspecified, and our search retrieved no matched verbal-tic evaluation of the 0915 Pro snapshot. Current model identity and the May application account therefore remain separate pieces of evidence.

Testing Seed 2.1 Pro through OpenCode in a June Zhihu review, \citet{zhihudoubao} found the presentation's content and layout less visibly AI-generated. The prompt explicitly requested short sentences, selected key points, and no promotional language. This favorable account reflects directed prompting on an unspecified June snapshot before the 0915 update.

\subsection{Moonshot AI}

Moonshot AI's API catalog identifies Kimi K3 as its strongest current model and records K2.5's discontinuation on 31 August 2026 \citep{kimi}. The official launch announcement separates initial product availability from the later open-weight release \citep{kimiannounce}.

Two Reddit roleplay posts assess K3's writing favorably \citep{kimiwriting1,kimiwriting2}. A July 28 post describes K3's prose as fresher and less formulaic after roughly two days of use under unspecified anti-slop prompts. An August post likewise prefers K3 over earlier Kimi generations. Personalized creative tasks of this kind help define conditions for testing both baseline formulaic prose and directed style adaptation.

A Chinese WeChat essay examines a targeted style-adaptation task \citep{wechatkimi}. Asked to study Daoerdeng's list-form essay about Qu Yuan and write a parallel piece about Zhuangzi, K3 produced a stylistic analysis and revisions that satisfied the author. We did not separately inspect the linked prompt, output, and reasoning record, and the reported length of that reasoning record must be kept separate from the finished essay.

\subsection{DeepSeek}

DeepSeek's live API identifiers map to V4-Pro-0813 and V4.1-Flash \citep{deepseekpricing,deepseekupdates}. That mapping separates the August Pro general-availability update from the smaller September Flash model.

A May 30 roleplay post using the official V4 Pro API reports recurring dialogue and praise across roughly 500 self-reported messages without supplying the full transcript \citep{deepseekpreview}. Because that session occurred during the May Preview period, its observations do not transfer to the August Pro-0813 snapshot.

A May Preview comparison on Zhihu describes a narrow rule-extraction query expanding into an elaborate plan that added rules absent from the source documents \citep{zhihudeepseek}. Even so, the author preferred Pro for complex multi-document synthesis, and a visible comment noted repeated \emph{oh wait} phrasing in OpenCode without recording the output layer or snapshot. Both points concern scope control and coding-workflow phrasing in an earlier deployment.

A September V2EX thread explicitly names V4.1 Flash and criticizes coined expressions, unusual syntax, and convoluted prose while praising several completed technical tasks \citep{v2exdeepseek}. Replies span agreement, alternative explanations, and favorable assessments. The thread documents readability concerns on a current release without providing an incidence rate, and user comments attributing the wording to distillation remain speculative.

An August Weibo post complains of repeated description and stacked modifiers in DeepSeek prose without naming a version, interface, or prompt \citep{weibodeepseek}. Published before Pro-0813, it illustrates a familiar reader complaint whose model-level source remains unresolved.

\subsection{Xiaomi}

MiMo-V2.6-Pro launched on 22 September 2026 \citep{mimorelease}. A September 27 technical post diagnosed repeated tool calls and documented an API service replacement on 25 September at 06:00 UTC+8 under unchanged model names \citep{mimorepetition}.

Xiaomi's report counts repetition when a single response contains identical tool names and normalized JSON arguments, and its reported rates vary by agent framework. Cross-turn repetition and near-identical arguments fall outside that metric. The disclosure supplies action-level evidence and an exact deployment boundary while leaving conversational catchphrases unmeasured.

Subsequent Hacker News discussion praised the disclosure without supplying an independent reproduction \citep{hnmimo}, and our retrieved set includes no current-release corpus of MiMo prose tics. Even so, the tool-call case clarifies service versioning and outcome definition.

\subsection{Synthesis across channels}

Retrieved sources describe recurrent marker words, strained technical phrasing, recycled dialogue, excessive explanation, and unwarranted validation alongside accounts of cleaner prose. Provider disclosures contribute audit definitions and procedures, benchmarks supply standardized tasks, and public forums capture everyday application settings. Convergence across channels justifies formal testing, yet their underlying measurements remain distinct.

Several accounts indicate that reading effort warrants its own evaluation measure. Dense wording in Astra and awkward terminology in V4.1 Flash concern how information is expressed rather than task failure \citep{hnastra,v2exdeepseek}. The Flash critic also praised completed technical work, and a September X comparison favored Gemini 3.8 Flash for coding and 3.1 Pro for advice \citep{xgeminineg}. Evaluations should therefore score task completion and prose readability separately against explicit audience and task expectations.

Directed prompts also complicate cross-model comparisons. Favorable reports on Seed presentations and Astra fiction used concision instructions or external writing skills \citep{zhihudoubao,wechatastro}, and K3 roleplay and imitation accounts relied on custom prompts \citep{kimiwriting1,wechatkimi}. Because those gains reflect model and prompt together, controlled evaluations should test both default output and adaptation to style instructions. Preference-training studies likewise offer plausible explanations for habitual agreement, whereas public user reports cannot isolate training mechanisms inside individual services.

Cross-platform retrieval requires tracking provenance. A YouTube review reposted on Bilibili belongs to the same source family as its original, and multiple posts quoting one release announcement share a single origin. Independent user accounts of a recurring pattern can guide sampling once their tasks, settings, and model attributions are preserved.

Direct evidence remains uneven across current releases. Some sources identify a current API snapshot or interface version, whereas others describe previews, retired models, or unspecified applications. Our catalog and ledger document this asymmetry directly without assigning a common score across the eight families.

\section{English and Chinese considerations}

English marker-word lists transfer poorly to Chinese. Chinese text analysis depends on explicit segmentation rules, programming discussions mix English identifiers with Chinese explanations, and discourse norms vary by register. Bilingual evaluations should retain original texts alongside translations and use raters familiar with both the language and the task.

On V2EX, the V4.1 Flash thread debates awkward Chinese terminology and separates task completion from readable explanation \citep{v2exdeepseek}. An earlier V2EX thread on DeepSeek-R1 instead tracks recurring openers in visible reasoning traces and separates them from final answers across a large self-reported trial count \citep{v2exr1}. Merging those two output layers would change what is measured.

A July Chinese V2EX thread on ChatGPT describes lengthy, document-like answers and shifting recommendations during follow-up turns on an unspecified backend \citep{v2exchatgpt}. Replies suggest prompt remedies and report mixed experiences without validating them. Those exchanges reflect reader concern over reading effort and conversational stability.

Chinese social posts also debate rhetorical templates. A September Xiaohongshu note attributes frequent contrast framing and Chinese connectors meaning \emph{even} and \emph{however} to ChatGPT without naming a backend \citep{xhstics}. On Weibo, one writer uses constructed examples to criticize contrasts that insert an unnecessary negation before linking loosely related claims \citep{weibocontrast}. A second Weibo writer notes that ordinary human prose uses the same construction and recommends judging whole passages \citep{weibocounter}. Because a substantive distinction and empty emphasis can share one grammatical form, surface matching requires contextual annotation.

Bilingual prompts should match communicative purpose rather than literal wording alone. Emotional support, factual correction, refusal, and technical explanation carry different expectations, and a difference between two prompt sets does not by itself establish a cultural explanation.

\section{A framework for future evaluation}
\label{sec:framework}

We outline a measurement protocol for empirical work rather than a validated composite index.

\subsection{Record the service and output layer}

Record the provider, exact model ID, snapshot, UTC timestamp, interface, system instructions, user prompt, history, reasoning setting, and active generation parameters. Mark defaults explicitly. For open weights, preserve the repository revision, quantization, inference software, and chat template. When an application hides its backend, retain the application label and date.

MiMo's in-place service replacement shows why collection timestamps matter \citep{mimorepetition}. Paired comparisons should hold task content constant and publish prompts. Separate default prompting from explicit style requests, and keep final answers distinct from reasoning traces, progress messages, and tool actions.

\subsection{Recurrence and contextual appropriateness}

Let \(N_{m,\ell,d}\) denote eligible responses for model \(m\), language \(\ell\), and domain \(d\). For pattern \(p\), response incidence is
\[
 I_{p,m,\ell,d} =
 \frac{\sum_{i=1}^{N_{m,\ell,d}}\mathbf{1}\{p\text{ occurs in response }i\}}
 {N_{m,\ell,d}}.
\]
Document the matcher, inflections, punctuation variants, and handling of repeated hits. Response incidence and occurrences per thousand words answer different questions.

Let \(A_{i,p}=1\) when at least one occurrence of pattern \(p\) in response \(i\) is judged unnecessary or inappropriate, and \(A_{i,p}=0\) otherwise. Contextually unwarranted incidence is
\[
 U_{p,m,\ell,d} =
 \frac{\sum_{i=1}^{N_{m,\ell,d}}\mathbf{1}\{p\text{ occurs in }i\}A_{i,p}}
 {N_{m,\ell,d}}.
\]
Annotators need examples of appropriate use alongside errors. Report inter-annotator agreement and unresolved cases. Estimating this second quantity requires contextual judgment beyond lexical matching.

Discover and test candidate patterns on separate samples, and validate lists learned from one model or genre elsewhere. Human baselines must also match register and purpose because an encyclopedia article and a support conversation follow different norms.

\subsection{Agreement, length, and dependence}

For factual tasks, record the initial answer, challenge, new evidence, and revision, and separate shifts toward truth from shifts away from truth. For advice tasks, assess the grounds for validation alongside the proposed actions. Code empathy, politeness, praise, and unsupported endorsement separately.

Blind model identity for human raters where feasible. Score readability, relevance, content faithfulness, and support independently. Comprehension or behavioral measures can supplement stated preference.

Longer responses create more opportunities for any pattern to appear, so report length and apply appropriate adjustments. State window sizes for lexical-diversity measures. Multi-turn tests should cluster uncertainty estimates by conversation, and paired tasks should resample tasks rather than individual sentences. Report denominators and run-level variation.

Test style-prompt interventions alongside accuracy and content coverage. Ban lists can alter surface words without changing judgments. Sampling settings should follow each service's supported parameters because identical numerical temperatures need not produce comparable randomness.

\subsection{Transparent measures}

An informative summary reports matched-pattern incidence, unwarranted use, local repetition, response length, relevant-content coverage, and task-appropriate agreement outcomes by language and task.

EQ-Bench's Slop Score publishes explicit weights for overused words, contrast constructions, and trigrams in creative writing and essays \citep{slop}, and Creative Writing v3 documents a word-and-phrase repetition measure \citep{eqcreative}. Both methods offer transparent starting points that keep their task-specific names and definitions, whereas conversational and Chinese-language settings require matched validation.

\section{Limitations and research priorities}

Retrieved sources differ in access depth and evidentiary strength. Some studies were available as full manuscripts or published articles, whereas others supplied publisher abstracts and structured summaries. Live API catalogs and public discussions also change, so the source ledger documents attribution and access without replacing an archived corpus.

Community coverage reflects who posts and which pages remain searchable. Roleplay and programming forums value different styles, and their templates, memories, anti-slop prompts, and conversation histories shape output. Shared complaints can motivate candidate tests even though they remain unsuitable for population estimates.

Direct evidence on the newest releases is uneven. Published studies apply stronger designs to earlier models, whereas current discussions supply narrower observations. Empirical work should prioritize paired sampling of documented current snapshots across short and extended conversations with separate final-answer and reasoning corpora. Contextual annotation in English and Chinese should include legitimate uses of flagged expressions. Longitudinal collection across service updates can then test whether recurrence persists or shifts. Studies should publish dated configurations, prompts, and output samples for independent checking.

\section{Conclusion}

Published research supports treating recurrent vocabulary and discourse structure as measurable aspects of writing quality. Controlled studies also show that persona interventions can alter accuracy and agreement, and that validating social advice can shift conflict judgments. Each finding applies to the models and conditions tested.

Our release catalog and dated discussion ledger show that model deployment outpaces comparable behavioral measurements in the reviewed evidence, even as several current-model reports describe clearer writing. Future comparisons should therefore test both complaints and favorable observations on matched tasks, measure recurrence in context, and preserve exact service and output-layer records. Such designs can determine where a pattern occurs, what it contributes to communication, and whether it alters understanding or judgment.

\section*{Source and manuscript availability}
Public source links appear in the bibliography and discussion ledger. The monthly data in Figure~\ref{fig:corpus} come from the publisher's Figure~1 source workbook. That workbook, the extracted monthly records, and the extraction and plotting scripts are retained with the manuscript files. Evidence descriptions remain tied to the cited source texts.

\appendix
\section{Discussion source ledger}
\label{app:ledger}

This register catalogs the qualitative discussion sources. Readable source text refers to the main post or specified comments rather than uninspected images or full conversation histories. Dates retain each source's displayed timestamp and mark UTC where needed. All entries were accessed for the 1 October 2026 snapshot.

\begingroup\small\setlength{\tabcolsep}{3pt}\renewcommand{\arraystretch}{1.17}
\begin{longtable}{@{}P{\dimexpr 0.203822\linewidth-3.66879pt\relax}P{\dimexpr 0.210191\linewidth-3.78344pt\relax}P{\dimexpr 0.286624\linewidth-5.15924pt\relax}P{\dimexpr 0.299363\linewidth-5.38854pt\relax}@{}}
\caption{Discussion provenance and limits. Each row is one source rather than an independent statistical sample.}\label{tab:ledger}\\
\toprule \textbf{Source and date} & \textbf{Model attribution} & \textbf{Readable observation} & \textbf{Principal limit}\\ \midrule
\endfirsthead
\multicolumn{4}{l}{\textbf{Table \thetable{}, continued}}\\ \toprule
\textbf{Source and date} & \textbf{Model attribution} & \textbf{Readable observation} & \textbf{Principal limit}\\ \midrule
\endhead
\midrule\multicolumn{4}{r}{Continued on next page}\\
\endfoot
\bottomrule
\endlastfoot
HN, 22 Sep 2026; \citealp{hnastra} & GPT-6 Astra & Main post reports denser prose and higher reading effort & Complete prompt and output sample absent; mechanism hypothesis excluded.\\
Tom's Guide, 25 Sep 2026; \citealp{caswell} & Opus 5.5; ChatGPT-6 UI label & Author's writing and everyday-task comparison & Personal trial; exact ChatGPT API and full sample absent.\\
GitHub, 3 Aug 2026; \citealp{githubclaude} & Earlier Opus, Sonnet, and Fable versions & Issue body, prompts, examples, and reproduction links & Report not rerun; mixed effort settings and categories.\\
Reddit, 1 May 2026; \citealp{redditgemini} & Gemini 3.1 & Main post describes lengthy explanations after correction & Product impression; claimed silent quantization unverified.\\
Reddit, 31 Aug 2026; \citealp{redditflash} & Flash 3.5, 3.6, and 3.7 & Main post compares writing and objectivity preferences & Title anticipates 3.8; no 3.8 test.\\
Cursor, 21--25 Sep 2026; \citealp{groktalk} & Grok 4.7 & Comments 27, 34, and 110 note cadence, metaphors, and repeated explanations & Single correlated release thread; application settings vary.\\
Signed report, 14 May 2026; \citealp{doubaotest} & Doubao application & Author's challenge reversal and casual apology example & Backend unspecified; predates current Pro snapshot.\\
Reddit, 28 Jul 2026; \citealp{kimiwriting1} & Kimi K3 & Main post praises writing and mentions anti-slop prompting & Roughly two days of use; comments not retrieved.\\
Reddit, 15 Aug 2026 UTC; \citealp{kimiwriting2} & Kimi K3; earlier Kimi models & Main post expresses preference for K3 prose & Subjective roleplay; comments not retrieved.\\
Reddit, 30 May 2026; \citealp{deepseekpreview} & V4 Pro Preview, official API & Main post reports recurring dialogue and flattering characterization & Self-reported long session; full response corpus absent.\\
V2EX, 12 Sep 2026; \citealp{v2exdeepseek} & V4.1 Flash & Main post and replies debate difficult prose and task results & Anecdotal; proposed distillation cause unverified.\\
V2EX, 14 Feb 2025; \citealp{v2exr1} & DeepSeek-R1 & Main post and replies note recurring reasoning openers & Separated from final answers; trial count self-reported.\\
V2EX, 17 Jul 2026; \citealp{v2exchatgpt} & ChatGPT Go application & Main post and replies describe length, framing, and shifting advice & Backend unidentified; prompting remedies unvalidated.\\
HN, 27 Sep 2026; \citealp{hnmimo} & MiMo-V2.6 technical disclosure & Single comment praises disclosure transparency & Independent linguistic or tool-call reproduction absent.\\
WeChat, 5 Sep 2026; \citealp{wechatastro} & GPT-6 Astra & Full article reports precise story diction and repetitive science continuation & Tasks and skill settings differ; screenshots uninspected.\\
YouTube, 3 Sep 2026 UTC; \citealp{berman} & GPT-6 Astra, early access & Automatic subtitles cover attribution at 0:00--0:45 and writing appraisal at 13:41--14:17 & Passages read without video review; text sample absent.\\
X, 23 Sep 2026 UTC; \citealp{xopus} & Opus 5 and 5.5 & Full post favors 5.5 clarity in writing trial & Prompt and output sample absent; quoted marketing kept separate.\\
Xiaohongshu, 23 Sep 2026; \citealp{xhsopus} & Opus 5.5 & Full note reports less AI-like style after one day of use & Writing task unspecified; images and comments uninspected.\\
LinkedIn, 28 Sep 2026 UTC; \citealp{linkedinevery} & Sonnet 5.5 and Opus 5.5 & Full team post reports readable prose and mixed task preferences & Team's own summary; complete test records absent.\\
X, 3 Sep 2026 UTC; \citealp{xgeminipos} & Gemini 3.8 & Full post praises practical, readable English & Flash designation omitted; prompts and outputs absent.\\
X, 16 Sep 2026 UTC; \citealp{xgeminineg} & Gemini 3.8 Flash; 3.1 Pro & Full post praises Flash coding and judges advice moralizing & Personal cross-task preference; conditions incomplete.\\
Zhihu, 23 Jun 2026; \citealp{zhihudoubao} & Seed 2.1 Pro in OpenCode & Full article gives favorable presentation assessment & June snapshot unspecified; prompt directs concise style.\\
WeChat, 25 Jul 2026; \citealp{wechatkimi} & Kimi K3 & Full article gives favorable style-imitation account & Linked prompt, output, and reasoning record not separately read.\\
Zhihu, 30 May 2026; \citealp{zhihudeepseek} & V4 Pro and Flash in Preview period & Full article and selected comments note scope expansion and mixed task preferences & Original dialogues and configuration absent; comments partial.\\
Weibo, 5 Aug 2026; \citealp{weibodeepseek} & DeepSeek, version unspecified & Full main post reports repeated description and modifiers & Predates Pro-0813; attached image uninspected.\\
Xiaohongshu, 9 Sep 2026; \citealp{xhstics} & ChatGPT, backend unspecified & Full note describes contrast framing and repeated connectors & Images and comments uninspected; exact model absent.\\
Weibo, 8 Sep 2026; \citealp{weibocontrast} & AI-writing discussion, model unspecified & Full main post criticizes unnecessary negation in contrast framing & Author-constructed examples; model test and comments absent.\\
Weibo, 5 Sep 2026; \citealp{weibocounter} & General writing discussion & Full main post notes ordinary human use of contrasts & Counterview on attribution; model test and comments absent.\\
\end{longtable}
\endgroup

\section{Platform coverage}
\label{app:coverage}

Table~\ref{tab:coverage} records the searches and reading completed for the cutoff snapshot. Coverage refers to publicly retrievable material under targeted queries and does not estimate the share of relevant posts read on a platform. Each claim is bounded by its source's access depth. Platform search results, article prose, comments, screenshots, and video subtitles remain distinct.

\begingroup\small\setlength{\tabcolsep}{3pt}\renewcommand{\arraystretch}{1.17}
\begin{longtable}{@{}P{\dimexpr 0.164557\linewidth-1.97468pt\relax}P{\dimexpr 0.386076\linewidth-4.63291pt\relax}P{\dimexpr 0.449367\linewidth-5.39241pt\relax}@{}}
\caption{Executed platform searches and access depth. Searches with no selected source remain documented as retrieval gaps.}\label{tab:coverage}\\
\toprule \textbf{Platform} & \textbf{Representative discovery and reading} & \textbf{Selection or access limit}\\ \midrule
\endfirsthead
\multicolumn{3}{l}{\textbf{Table \thetable{}, continued}}\\ \toprule
\textbf{Platform} & \textbf{Representative discovery and reading} & \textbf{Selection or access limit}\\ \midrule
\endhead
\midrule\multicolumn{3}{r}{Continued on next page}\\
\endfoot
\bottomrule
\endlastfoot
Weibo & Current Doubao, K3, and V4 model names paired with stock phrases, flattery, repetition, and writing; selected full main posts & Unspecified-model complaints and rhetorical debate retrieved; no matched 0915, Pro-0813, or V4.1-Flash tic test selected.\\
Zhihu & Current-model and general verbosity queries; full V4 comparison, selected V4 comments, and Seed 2.1 review & Selected accounts concern earlier snapshots; some pages inaccessible and comment reading partial.\\
WeChat public accounts & Topic and exact-title searches; full Astra and K3 article prose & Narrative accounts selected; output screenshots and external records uninspected; access varies by request.\\
Xiaohongshu & Initial search pages for GPT catchphrases and GPT-6 Astra writing; five full note texts & First search returned 19 notes and second returned 17 notes plus 3 ads; selected notes cited without reading images or comments.\\
Bilibili & Grok 4.7 writing, DeepSeek catchphrases, and GPT-6 writing; video metadata and descriptions & Selected Astra video returned empty subtitles; no video-based style claim drawn; reposts deduplicated against originals.\\
Douyin & Doubao stock-phrase and catchphrase queries in web index and native video search & Indexed aggregation pages only and native request failed after a documented retry; no original post selected.\\
Kuaishou & Doubao and AI catchphrase queries via web discovery & Landing and general pages returned; no relevant firsthand account selected.\\
X / Twitter & Current Western model names paired with verbosity, jargon, sycophancy, and writing style; platform searches and post records & Selected full posts supply dated impressions with sparse task records.\\
YouTube & Astra, Opus 5.5, and Grok 4.7 writing and review queries; video metadata and selected automatic-subtitle passages & Video frames uninspected; Berman passage cited with timestamps.\\
LinkedIn & Astra writing and verbosity, Opus writing trials, and Gemini style queries; full Every team post & Team report selected; other hits largely repeat release announcements.\\
Threads & Opus 5.5 writing in web index; platform searches for Opus 5.5 and Gemini 3.1 & Platform searches returned empty lists and indexed hits showed profiles or unrelated pages; no source selected.\\
Mastodon & GPT-6 and Claude writing queries across selected public instances & News, topics, and general remarks returned without version-qualified firsthand tests; coverage is instance-limited.\\
Facebook & ChatGPT formulaic writing and sycophancy query & General application, memory, and story-continuity posts returned; no qualifying current-model source selected.\\
Instagram & ChatGPT repetitive writing and sycophancy query; one public caption and visible comments read & Portuguese copywriting post lies outside focal English and Chinese model evidence; video unviewed.\\
TikTok & ChatGPT verbal-tic queries in web discovery and native video search & Native request returned an error and web discovery yielded no qualifying current-model source.\\
Quora & ChatGPT repetition and verbosity queries; generic answer page read & General opinions lacked model versions; no current-version account selected.\\
Telegram & GPT-6 Astra and Opus 5.5 writing query in public channel index & Mainly bots and promotional feeds; no qualifying source selected.\\
Reddit & Model-specific style and repetition queries; selected full main posts & Roleplay and Gemini threads selected; comment access recorded per ledger entry.\\
Hacker News & Astra writing and MiMo disclosure queries; main post and specified comment & Firsthand impression and disclosure comment selected; independent current-model corpus absent.\\
V2EX & DeepSeek and ChatGPT language-pattern queries; full threads retrieved & Current Flash thread kept separate from earlier reasoning and product-level discussions.\\
Cursor forum & Grok 4.7 release thread and specified comments & Multiple observations drawn from one correlated thread.\\
GitHub & Claude verbosity issue with prompts and reproduction links & Earlier versions and mixed settings; reproduction not rerun.\\
\end{longtable}
\endgroup

Representative native queries paired Chinese terms for GPT catchphrases, GPT-6 Astra writing, and Western model names with verbosity or writing style. Expanding coverage on Weibo and Zhihu used 41 domain-targeted searches, and one search queried both platforms jointly. Search counts record work performed rather than eligible evidence counts.

\section{Retrieval and comparison notes}

Representative research queries included \emph{LLM formulaic writing patterns academic study}, \emph{LLM sycophancy vs politeness research paper}, \emph{delve excess vocabulary large language models scientific abstracts study}, and \emph{EQ-Bench slop score creative writing benchmark model methodology 2026}. Developer and model queries and backward reference tracing followed.

Direct discussion claims required readable primary texts, and items available only as search snippets were excluded from the ledger. Revised posts and reposts were grouped into shared source families when assessing independence.

Repeat reviews should inspect live release pages and API catalogs first, then preserve the exact models and test conditions in each study or post. Because updates and edits can change samples and conclusions, source revision, retrieval date, and model release date belong in separate fields.

\end{document}